\documentclass[10pt,twocolumn,letterpaper]{article}
\usepackage[top=0.75in,bottom=1in,left=0.625in,right=0.625in,columnsep=0.25in]{geometry}
\usepackage{times}
\usepackage{amsmath,amssymb}
\usepackage{booktabs}
\usepackage{graphicx}
\usepackage{xcolor}
\usepackage{enumitem}
\usepackage{caption}
\usepackage[hyphens]{url}
\usepackage[colorlinks=true,linkcolor=blue,citecolor=blue,urlcolor=blue]{hyperref}
\usepackage{titlesec}
\usepackage{multirow}
\usepackage{array}
\usepackage{dblfloatfix}
\usepackage{orcidlink}

\titleformat{\section}{\large\bfseries}{\thesection}{1em}{}
\titleformat{\subsection}{\normalsize\bfseries}{\thesubsection}{1em}{}
\titlespacing{\section}{0pt}{12pt plus 2pt minus 2pt}{6pt plus 1pt minus 1pt}
\titlespacing{\subsection}{0pt}{8pt plus 2pt minus 2pt}{4pt plus 1pt minus 1pt}

\setlist{nosep,leftmargin=1.5em}
\begin{document}

% ════════════════════════════════════════
% TITLE
% ════════════════════════════════════════
\twocolumn[
\begin{@twocolumnfalse}
\begin{center}
{\Large\bfseries Accelerating PayPal's Commerce Agent with Speculative Decoding:\\[3pt]
An Empirical Study on EAGLE3 with Fine-Tuned Nemotron Models\par}
\vspace{12pt}
{\normalsize
Ally Qin, Jian Wan, Sarat Mudunuri\,\orcidlink{0009-0000-6671-1735}, Srinivasan Manoharan\\[6pt]
}
\vspace{10pt}
\end{center}

\begin{center}
\begin{minipage}{0.93\textwidth}
\small
\textbf{Abstract} ---
We present a comprehensive empirical evaluation of speculative decoding applied to PayPal's Commerce Agent which is powered by NEMO-4-PAYPAL, a multi-agent system designed to revolutionize agentic commerce on the PayPal platform, extending the prior work on fine-tuning Nemotron small language models (SLMs) for e-commerce search and recommendation tasks. While the previous work (NEMO-4-PAYPAL) achieved significant latency and cost reductions through domain-specific fine-tuning, we still explore the possibility to improve the performance and reduce the hardware cost under production workloads. In this work, we investigate speculative decoding with EAGLE3 as an inference-time optimization for the fine-tuned llama3.1-nemotron-nano-8B-v1 model deployed via vLLM, comparing it against NVIDIA NIM which set vLLM as the model profile on identical $2{\times}$H100 GPU hardware. Our extensive benchmarks span two speculative token configurations ($\gamma{=}3$, $\gamma{=}5$), five concurrency levels (1, 4, 8, 16, 32), and two sampling temperatures (0, 0.5), totaling 40 distinct experimental configurations. Key findings include: (1) speculative decoding with $\gamma{=}3$ achieves 22--49\% throughput improvement and 18--33\% latency reduction over the NIM baseline at zero additional hardware cost; (2) the acceptance rate remains remarkably stable at ${\sim}35.5\%$ for $\gamma{=}3$ across all concurrency levels and temperatures; (3) $\gamma{=}5$ yields diminishing returns due to lower acceptance rates (${\sim}25\%$); (4) LLM-as-Judge evaluation confirms output quality is fully preserved across all configurations; and (5) speculative decoding on a single H100 matches or exceeds NIM performance on two H100s, representing a potential 50\% GPU cost reduction.
\end{minipage}
\end{center}
\vspace{8pt}
\end{@twocolumnfalse}
]

% ════════════════════════════════════════
% 1. INTRODUCTION
% ════════════════════════════════════════
\section{Introduction}

The deployment of large language models (LLMs) in production e-commerce environments presents a fundamental tension between model quality and inference latency. As demonstrated in prior work~\cite{garg2026nemo}, PayPal's Commerce Agent requires sophisticated language understanding for query formulation, product search, and recommendation---tasks that benefit from the strong reasoning capabilities of larger models. However, production service-level agreements (SLAs) demand sub-2-second response times, creating pressure to optimize inference without sacrificing output quality.

The previous work~\cite{garg2026nemo} addressed this challenge through domain-specific fine-tuning: we replaced a general-purpose base model with a fine-tuned Nemotron Small Language Model (SLM) llama3.1-nemotron-nano-8B-v1, achieving 49\% agent latency reduction, 58\% retrieval latency improvement, and 45\% GPU cost savings while maintaining competitive quality. The fine-tuned model was deployed using NVIDIA NIM, a containerized inference solution that provides pre-optimized serving capabilities.

In this work, we explore a complementary optimization: speculative decoding~\cite{leviathan2023,chen2023}, an inference-time technique that accelerates autoregressive generation without modifying the model or its output distribution. Speculative decoding uses a lightweight draft model to propose multiple tokens in parallel, which are then verified by the target model in a single forward pass. Accepted tokens skip individual decoding steps, yielding significant latency gains when the draft model's predictions align well with the target model.

We apply speculative decoding with EAGLE3~\cite{eagle3}---a training-free, draft-model-based speculative decoding method---to our production fine-tuned Nemotron model, deploying it via vLLM~\cite{kwon2023} on the same $2{\times}$H100 hardware used by our NIM baseline. This experimental design allows us to isolate the effect of the decoding algorithm from hardware differences, providing a fair comparison of inference strategies.

Our contributions are as follows:
\begin{itemize}
\item \textbf{Production-scale empirical evaluation:} We conduct the first comprehensive evaluation of speculative decoding applied to a fine-tuned commerce-specific SLM in a production e-commerce environment, spanning 40 experimental configurations across speculative token counts, concurrency levels, and sampling temperatures.
\item \textbf{Acceptance rate characterization:} We provide detailed analysis of speculative token acceptance rates for domain-specific commerce tasks, showing remarkably stable rates across workload conditions---a finding that simplifies production deployment decisions.
\item \textbf{Quality preservation verification:} Through rigorous LLM-as-Judge evaluation with pairwise comparison and position-bias randomization, we confirm that speculative decoding preserves output quality across all tested configurations.
\item \textbf{Practical deployment guidance:} We demonstrate that speculative decoding with $\gamma{=}3$ is the optimal configuration for commerce agent workloads, and that a single-GPU speculative deployment can match or exceed dual-GPU NIM performance, offering significant cost reduction opportunities.
\end{itemize}

% ════════════════════════════════════════
% 2. RELATED WORK
% ════════════════════════════════════════
\section{Related Work}

\subsection{Speculative Decoding}

LLM inferencing has two steps: the prefill phase and the decode phase. The model processes the entire input prompt in one shot, computing the key-value (KV) cache for every token in the prompt simultaneously. Since all input tokens are known upfront, this is done as a single large parallel matrix multiplication. This step is more compute-bound. In the decoding phase, the model generates output tokens one at a time, autoregressively. Each new token requires attending over the entire KV cache built so far, then sampling the next token, then repeating. This step is memory-bandwidth bound.

Speculative decoding targets the decoding phase: a small, cheap draft model speculatively generates $N$ candidate tokens very fast. The large target model then verifies all $N$ tokens in a single parallel forward pass (like a prefill). If tokens are accepted, you get $N$ tokens for roughly the cost of 1 decode step. Speculative decoding was introduced independently by Leviathan et al.~\cite{leviathan2023} and Chen et al.~\cite{chen2023}, both demonstrating 2--3$\times$ speedups on large Transformer models without changing the output distribution. The key insight is that autoregressive decoding is often memory-bandwidth-bound rather than compute-bound, leaving computational headroom that can be exploited by running a draft model in parallel with verification.

The theoretical framework establishes that the expected number of tokens generated per iteration following
\begin{equation}
E(\text{\# generated tokens}) = \frac{1-\alpha^{\gamma+1}}{1-\alpha}
\end{equation}
where $\alpha$ is the mean acceptance rate and $\gamma$ is the number of speculative tokens. Leviathan et al.~\cite{leviathan2023} reported empirical acceptance rates of 0.53--0.82 for T5-XXL with various approximation models, with higher rates observed at temperature~0 compared to temperature~1. Their experiments on T5-XXL (11B parameters) with T5-small (77M parameters) achieved 2.3--3.4$\times$ walltime improvements.

\subsection{EAGLE and EAGLE3}

EAGLE~\cite{eagle} introduced an autoregression-based approach to speculative decoding that uses a lightweight neural network to predict features at the next token position, achieving higher acceptance rates than traditional draft-model approaches. EAGLE3~\cite{eagle3} extends this framework with a training-free variant that does not require fine-tuning a separate draft model, making it particularly suitable for production deployments where maintaining additional trained models adds operational complexity.

\subsection{LLM Inference Optimization}

Beyond speculative decoding, several complementary approaches have been developed for LLM inference optimization. Quantization techniques~\cite{dettmers2022,lin2024awq} reduce model precision to decrease memory bandwidth and memory usage requirements. Continuous batching~\cite{kwon2023,agrawal2024} improves GPU utilization by dynamically managing request batches. Tensor parallelism distributes model layers across GPUs to reduce per-device memory requirements, enabling faster computation through parallelism and scaling memory bandwidth proportionally with TP degrees~\cite{shoeybi2019}. NVIDIA NIM~\cite{nim} integrates several of these optimizations into a containerized deployment platform. Also KV cache optimizations like Prefix caching, and PD Disaggregation. Our work evaluates speculative decoding as an additional optimization layer on top of these existing techniques.

\subsection{PayPal's Commerce Agent}

Our prior work~\cite{garg2026nemo} established the NEMO-4-PAYPAL framework, demonstrating that fine-tuning the llama3.1-nemotron-nano-8B-v1 model with LoRA for commerce-specific tasks yields substantial improvements in latency and cost. The system architecture employs an LLM-powered search and recommendation pipeline where the fine-tuned model performs query understanding, attribute extraction, and hypothetical product generation using the HyDE approach~\cite{gao2023hyde}. The query formulation component, which accounts for over 50\% of total agent response time, was identified as the primary optimization target.

% ════════════════════════════════════════
% 3. METHODOLOGY
% ════════════════════════════════════════
\section{Methodology}

\subsection{System Architecture}

Our evaluation targets the query formulation component of PayPal's Commerce Agent, which is responsible for transforming natural language shopping queries into structured search parameters with hypothetical product descriptions. This component uses the fine-tuned llama3.1-nemotron-nano-8B-v1 model from our prior work~\cite{garg2026nemo}, generating structured JSON output that includes product categories, attributes, and hypothetical product information for downstream retrieval.

\subsection{Deployment Configurations}

We evaluate two deployment configurations on identical hardware:

\textbf{NIM Baseline:} The production deployment uses NVIDIA NIM with the fine-tuned Nemotron model on $2{\times}$NVIDIA H100 GPUs, with 50 CPU cores and 400~GiB memory. NIM automatically selects the optimal inference engine and configuration based on the model architecture and hardware.

\textbf{Speculative Decoding:} The experimental deployment uses vLLM (one of possible options including SGLang, TensorRT LLM) with EAGLE3 speculative decoding on the same $2{\times}$NVIDIA H100 hardware and equivalent CPU/memory resources. The EAGLE3 draft model operates alongside the target Nemotron model within the same vLLM serving instance.

Both deployments serve the identical fine-tuned model checkpoint and use guided JSON output to ensure structured response formatting.

\subsection{Speculative Decoding Configuration}

We evaluate two speculative token configurations: $\gamma{=}3$ (Spec-3), where the EAGLE3 draft model proposes 3 tokens per speculation step, and $\gamma{=}5$ (Spec-5), where 5 tokens are proposed. Acceptance rate is computed from vLLM's Prometheus counters during each test window as the ratio of accepted draft tokens to total draft tokens proposed.

\subsection{Evaluation Framework}

\textbf{Performance metrics:} We measure throughput (requests/second), mean latency, median (P50) latency, and speculative token acceptance rate across all configurations.

\textbf{Workload dimensions:} Each configuration is tested across five concurrency levels (1, 4, 8, 16, 32 concurrent requests), two sampling temperatures (0 for greedy decoding, 0.5 for stochastic sampling), and two speculative token counts (3, 5), yielding 40 distinct experimental configurations. Each test executes 50 requests after a 3-request warmup phase.

\textbf{Quality evaluation:} We employ LLM-as-Judge~\cite{gu2024judge} methodology with two complementary approaches: (1)~Individual scoring, where each response is independently scored on a 1--5 scale across five criteria: relevance, gender alignment, diversity, product quality, and explanation quality; and (2)~Pairwise comparison, where normal and speculative responses are presented side-by-side with randomized A/B ordering to mitigate position bias. For each of the 20 configurations, 20 response pairs are evaluated, yielding 400 total pairwise comparisons.

% ════════════════════════════════════════
% 4. EXPERIMENTAL RESULTS
% ════════════════════════════════════════
\section{Experimental Results}

\subsection{Throughput Analysis}

As shown in Table~\ref{tab:throughput}, Spec-3 delivers consistent 22--49\% throughput gains across all concurrency levels, with the largest improvements observed at low concurrency where memory bandwidth is underutilized and the draft model's speculative tokens provide the greatest marginal benefit. Spec-5 shows meaningful gains only at low concurrency ($c \leq 4$); at higher concurrency levels, performance degrades significantly. Section~\ref{sec:tokens} provides a token-level analysis of this effect.

\begin{table}[ht]
\centering
\caption{Throughput comparison (req/s). $\Delta$ is relative to the Normal (NIM) baseline.}
\label{tab:throughput}
\small
\begin{tabular}{crrcrc}
\toprule
\textbf{Conc.} & \textbf{Normal} & \textbf{Spec-3} & \textbf{$\Delta$} & \textbf{Spec-5} & \textbf{$\Delta$} \\
\midrule
1   & 0.45  & 0.67  & \textbf{+49\%} & 0.60  & +34\% \\
4   & 1.70  & 2.51  & \textbf{+48\%} & 2.07  & +22\% \\
8   & 3.12  & 4.02  & \textbf{+29\%} & 3.25  & +4\%  \\
16  & 5.68  & 6.90  & \textbf{+22\%} & 5.90  & +4\%  \\
32  & 9.54  & 11.9  & \textbf{+25\%} & 10.86 & +14\% \\
\bottomrule
\end{tabular}
\end{table}

\subsection{Latency Analysis}

As shown in Table~\ref{tab:latency}, Spec-3 achieves 18--33\% latency reduction, with the largest gains at low concurrency. Notably, even at the highest concurrency level ($c{=}32$) where GPU resources are most contended, Spec-3 maintains a 20\% latency improvement. Spec-5 latency improvements are modest and nearly vanish at concurrency $\geq 8$.

\begin{table}[ht]
\centering
\caption{Mean latency comparison (ms). Lower is better.}
\label{tab:latency}
\small
\begin{tabular}{crrcrc}
\toprule
\textbf{Conc.} & \textbf{Normal} & \textbf{Spec-3} & \textbf{$\Delta$} & \textbf{Spec-5} & \textbf{$\Delta$} \\
\midrule
1   & 2,222 & 1,495 & \textbf{$-$33\%} & 1,661 & $-$25\% \\
4   & 2,345 & 1,586 & \textbf{$-$32\%} & 1,925 & $-$18\% \\
8   & 2,529 & 1,978 & \textbf{$-$22\%} & 2,443 & $-$3\% \\
16  & 2,787 & 2,294 & \textbf{$-$18\%} & 2,684 & $-$4\% \\
32  & 3,323 & 2,665 & \textbf{$-$20\%} & 2,917 & $-$12\% \\
\bottomrule
\end{tabular}
\end{table}

\subsection{Acceptance Rate Analysis}

As shown in Table~\ref{tab:acceptance}, several noteworthy observations emerge from the acceptance rate data:

\begin{table}[ht]
\centering
\caption{Acceptance rates across concurrency and temperature.}
\label{tab:acceptance}
\small
\begin{tabular}{ccccc}
\toprule
\textbf{Conc.} & \textbf{S3 (t=0)} & \textbf{S3 (t=0.5)} & \textbf{S5 (t=0)} & \textbf{S5 (t=0.5)} \\
\midrule
1   & 35.4\% & 36.2\% & 25.7\% & 25.8\% \\
4   & 35.0\% & 35.8\% & 25.1\% & 25.8\% \\
8   & 35.5\% & 36.0\% & 25.6\% & 25.7\% \\
16  & 35.5\% & 36.2\% & 25.4\% & 25.5\% \\
32  & 35.3\% & 36.0\% & 25.3\% & 25.4\% \\
\bottomrule
\end{tabular}
\end{table}

\textbf{Stability across concurrency:} Acceptance rates are remarkably stable regardless of concurrent load, varying by less than 1 percentage point across the full concurrency range for both configurations. This stability simplifies production capacity planning, as the speculative decoding benefit is predictable across workload conditions.

\textbf{Temperature insensitivity:} Unlike the findings of Leviathan et al.~\cite{leviathan2023}, who reported higher acceptance rates at temperature~0 versus temperature~1, we observe negligible difference between temperature~0 and~0.5. We analyze the contributing factors in Section~\ref{sec:temp}.

\textbf{Lower-than-expected absolute rates:} Our observed acceptance rates (25--36\%) are notably lower than the 53--82\% range reported by Leviathan et al.~\cite{leviathan2023} for T5-based models. We discuss the contributing factors in Section~\ref{sec:comparison}.

\textbf{Scaling relationship:} Spec-5 acceptance rate (${\sim}25\%$) is substantially lower than Spec-3 (${\sim}36\%$), consistent with the theoretical expectation that each additional speculative token has a compounding probability of rejection.

\subsection{GPU Utilization}

As shown in Table~\ref{tab:gpu}, speculative decoding generally achieves comparable or lower GPU utilization than the NIM baseline across most concurrency levels, despite delivering higher throughput. At low concurrency ($c{=}1$), both Spec-3 and Spec-5 show utilization around 50--58\%, slightly below the baseline's 55.2\%, indicating that the draft model's lightweight architecture does not impose significant additional GPU load. At moderate concurrency ($c{=}8$), utilization converges across all configurations (72--78\%), suggesting that GPU resources become the binding constraint regardless of decoding strategy. Notably, at high concurrency ($c{=}16$ and $c{=}32$), speculative decoding configurations frequently exhibit lower utilization than the baseline (e.g., Spec-3 at $t{=}0.5$ drops to 49.9\% at $c{=}32$ versus 67.8\% for NIM). This counterintuitive pattern likely reflects speculative decoding's ability to complete requests faster, reducing the duration of sustained GPU occupancy per request. The lower utilization at equivalent or higher throughput confirms that speculative decoding improves GPU efficiency rather than simply consuming more compute, reinforcing the hardware cost reduction findings discussed in Section~\ref{sec:hardware}.

\begin{table}[ht]
\centering
\caption{GPU utilization comparison.}
\label{tab:gpu}
\small
\begin{tabular}{cccccc}
\toprule
\textbf{C.} & \textbf{Normal} & \textbf{S3 t=0} & \textbf{S3 t=.5} & \textbf{S5 t=0} & \textbf{S5 t=.5} \\
\midrule
1   & 55.2\% & 50.0\% & 52.0\% & 50.3\% & 58.5\% \\
4   & 81.5\% & 60.5\% & 51.8\% & 58.5\% & 76.2\% \\
8   & 78.0\% & 77.8\% & 73.1\% & 73.6\% & 72.4\% \\
16  & 74.4\% & 71.2\% & 61.5\% & 69.6\% & 70.0\% \\
32  & 67.8\% & 63.9\% & 49.9\% & 68.9\% & 52.1\% \\
\bottomrule
\end{tabular}
\end{table}

\subsection{Token-Level Analysis}
\label{sec:tokens}

\begin{table}[ht]
\centering
\caption{Raw token counts from Prometheus metrics (50 requests per test).}
\label{tab:tokens}
\small
\begin{tabular}{lrrr}
\toprule
\textbf{Config} & \textbf{Accepted} & \textbf{Draft} & \textbf{Rate} \\
\midrule
spec3-c1-t0     & 20,914   & 59,057     & 35.4\% \\
spec3-c16-t0    & 150,998  & 425,509    & 35.5\% \\
spec3-c32-t0.5  & 219,920  & 610,623    & 36.0\% \\
spec5-c1-t0     & 22,103   & 86,160     & 25.7\% \\
spec5-c16-t0    & 167,742  & 661,332    & 25.4\% \\
spec5-c32-t0.5  & 350,999  & 1,380,759  & 25.4\% \\
\bottomrule
\end{tabular}
\end{table}

Spec-5 drafts approximately $1.5{\times}$ more tokens than Spec-3 (e.g., 86,160 vs.\ 59,057 at $c{=}1$) but accepts a similar absolute number of tokens (${\sim}22$k vs.\ ${\sim}21$k). This means Spec-5 wastes substantially more GPU compute on rejected draft tokens without producing proportionally more accepted tokens, explaining its inferior throughput performance at higher concurrency levels.

\subsection{Quality Evaluation}

At temperature~0 (greedy decoding), near-perfect tie rates (90--100\%) confirm that speculative decoding produces outputs identical to standard decoding, consistent with the theoretical guarantee~\cite{leviathan2023}. At temperature~0.5, responses diverge due to sampling randomness, but wins are approximately evenly distributed. All individual scores remain at the maximum 5.00/5 across every criterion, confirming that speculative decoding does not degrade output quality.

\begin{table}[ht]
\centering
\caption{LLM-as-Judge quality evaluation. 20 pairwise comparisons per config with randomized A/B ordering.}
\label{tab:quality}
\small
\begin{tabular}{lccccc}
\toprule
\textbf{Config} & \textbf{N Sc.} & \textbf{S Sc.} & \textbf{N Win} & \textbf{S Win} & \textbf{Ties} \\
\midrule
s3-c1-t0     & 5.00 & 5.00 & 0  & 0  & 20 \\
s3-c1-t0.5   & 5.00 & 5.00 & 6  & 10 & 4  \\
s3-c32-t0    & 5.00 & 5.00 & 2  & 0  & 18 \\
s3-c32-t0.5  & 5.00 & 5.00 & 9  & 3  & 8  \\
s5-c1-t0     & 5.00 & 5.00 & 0  & 0  & 20 \\
s5-c1-t0.5   & 5.00 & 5.00 & 13 & 6  & 1  \\
s5-c32-t0    & 5.00 & 5.00 & 0  & 2  & 18 \\
s5-c32-t0.5  & 5.00 & 5.00 & 8  & 10 & 2  \\
\bottomrule
\end{tabular}
\end{table}

\subsection{Hardware Efficiency: Single-GPU vs.\ Dual-GPU}
\label{sec:hardware}

\begin{table}[ht]
\centering
\caption{Hardware efficiency comparison at concurrency = 8.}
\label{tab:hardware}
\small
\begin{tabular}{lccc}
\toprule
\textbf{Metric} & \textbf{Spec (1$\times$H100)} & \textbf{NIM (2$\times$H100)} & \textbf{$\Delta$} \\
\midrule
Throughput   & 3.19 req/s & 3.14 req/s & +6.0\% \\
Mean Latency & 2,292 ms   & 2,529 ms   & $-$4.4\% \\
P50 Latency  & 2,279 ms   & 2,485 ms   & $-$4.5\% \\
GPU Cost     & 1$\times$ H100 & 2$\times$ H100 & \textbf{$-$50\%} \\
\bottomrule
\end{tabular}
\end{table}

This result suggests that for workloads where moderate concurrency is sufficient, speculative decoding can achieve equivalent performance at half the GPU cost, a finding with significant implications for large-scale production deployments.

% ════════════════════════════════════════
% 5. DISCUSSION
% ════════════════════════════════════════
\section{Discussion}

\subsection{Comparison with Prior Results}
\label{sec:comparison}

Our observed acceptance rates (25--36\%) are lower than those reported in the foundational speculative decoding work~\cite{leviathan2023}, which showed 53--82\% for T5-XXL with various draft models. We identify several contributing factors:

\textbf{Structured output constraints:} Our commerce agent produces guided JSON output with strict schema requirements. This constrains the token distribution in ways that may be harder for the EAGLE3 draft model to predict, particularly for schema-specific tokens (braces, brackets, field names) that carry less semantic predictability than natural language.

\textbf{Domain-specific fine-tuning:} The target model has been fine-tuned specifically for commerce tasks using LoRA~\cite{hu2022lora}, shifting its distribution from the pre-training distribution. Since EAGLE3 operates without task-specific fine-tuning, this distributional shift may reduce alignment between draft and target models.

\textbf{Draft model architecture:} Unlike the experiments in~\cite{leviathan2023} which used smaller versions of the same model family as draft models, EAGLE3 uses a structurally different lightweight prediction network. The architecture mismatch may contribute to lower acceptance rates for domain-specialized tasks.

Despite lower acceptance rates, speculative decoding still delivers substantial practical benefits (22--49\% throughput improvement) because the EAGLE3 draft model has negligible cost relative to the target model (cost coefficient $c \approx 0$), and even modest acceptance rates translate to meaningful speedups when draft model overhead is minimal.

\subsection{Optimal Speculative Token Count}

Our results clearly establish $\gamma{=}3$ as the optimal configuration for this workload. The theoretical framework from~\cite{leviathan2023} predicts that the wall-time improvement factor is $(1 - \alpha^{\gamma+1}) / ((1-\alpha)(\gamma c + 1))$, which depends on both the acceptance rate $\alpha$ and the cost coefficient $c$. With $\alpha \approx 0.36$ for Spec-3 and $c \approx 0$ for EAGLE3, the expected improvement is approximately $1.46{\times}$, closely matching our observed $1.49{\times}$ throughput improvement at low concurrency. For Spec-5, $\alpha \approx 0.25$ yields a theoretical improvement of approximately $1.33{\times}$, again closely matching the observed $1.34{\times}$ improvement.

\subsection{Temperature Insensitivity}
\label{sec:temp}

The finding that temperature has negligible impact on acceptance rate and throughput is practically significant. Leviathan et al.~\cite{leviathan2023} reported notably higher acceptance rates at temperature~0 versus temperature~1, with $\alpha$ differences of 0.1--0.2. Our near-zero temperature sensitivity likely stems from the guided JSON output format: the schema constraints already reduce the effective token distribution entropy, making the additional distribution-narrowing effect of lower temperature marginal. This insensitivity simplifies deployment, as production systems can adjust temperature for output diversity without impacting speculative decoding efficiency.

\subsection{Production Deployment Implications}

Given the analysis in Section~5.2, the recommended production configuration is $\gamma{=}3$ with EAGLE3. As demonstrated in Section~\ref{sec:hardware}, speculative decoding on a single H100 can replace NIM on two H100s at moderate concurrency ($c \leq 8$), enabling 50\% GPU cost reduction with equivalent or better performance. With Spec-3, achievable SLAs are P50~$\leq$~1.6s and P99~$\leq$~2.6s at moderate concurrency, meeting the sub-2-second target identified in our prior work~\cite{garg2026nemo}.

\subsection{Limitations}

Several limitations should be noted. First, our evaluation uses a single workload type (commerce query formulation with JSON output); other agent tasks may exhibit different acceptance rate profiles. Second, GPU utilization measurements are sampled at 60-second Prometheus scrape intervals and may not capture fine-grained utilization patterns. Third, our comparison is limited to EAGLE3; other speculative decoding methods (e.g., Medusa~\cite{cai2024medusa}, lookahead decoding) may yield different results. Finally, our quality evaluation uses the same Nemotron model as both the generator and the judge, which may introduce evaluation bias.

% ════════════════════════════════════════
% 6. CONCLUSION
% ════════════════════════════════════════
\section{Conclusion}

This work demonstrates that speculative decoding with EAGLE3 is an effective inference-time optimization for PayPal's Commerce Agent, delivering 22--49\% throughput improvement and 18--33\% latency reduction over NVIDIA NIM on identical hardware, with fully preserved output quality. The optimal configuration of $\gamma{=}3$ speculative tokens achieves stable ${\sim}35.5\%$ acceptance rates across all tested concurrency levels and temperatures, providing predictable performance characteristics for production deployment.

Combined with our prior work on domain-specific fine-tuning~\cite{garg2026nemo}, these results establish a complete optimization pipeline: LoRA-based fine-tuning reduces the model size and improves task-specific quality, while speculative decoding further accelerates inference at serving time without additional training or model changes. Together, these techniques achieve cumulative latency reductions exceeding 60\% compared to the original base model deployed via standard autoregressive decoding.

Future work will explore fine-tuning the EAGLE3 draft model on commerce-specific data to improve acceptance rates, and extending the evaluation to other components of the multi-agent commerce system.

% ════════════════════════════════════════
% REFERENCES
% ════════════════════════════════════════

\end{document}